\documentclass[conference]{IEEEtran}

\usepackage{cite}
\usepackage{amsmath,amssymb,amsfonts}
\usepackage{graphicx}
\usepackage{textcomp}
\usepackage{booktabs}
\usepackage{url}
\usepackage{tikz}
\usetikzlibrary{decorations.pathreplacing}
\graphicspath{​{./}{../figures/}}

\def\BibTeX{{\rm B\kern-.05em{\sc i\kern-.025em b}\kern-.08em
    T\kern-.1667em\lower.7ex\hbox{E}\kern-.125emX}}

\begin{document}

\title{Nested Byte-Level Vocabularies Are Cheap to Deploy and\\
Expensive to Share: A Pre-Registered Negative Result}

\author{\IEEEauthorblockN{Christos Koutsiaris}
\IEEEauthorblockA{\textit{Cloud ERP -- UX Foundation} \\
\textit{SAP P\&E}\\
christos.koutsiaris@sap.com}
}

\maketitle

\begin{abstract}
A byte-level BPE tokenizer is an ordered list of merge rules, so applying only a prefix of them
yields a coarser vocabulary whose token identifiers are exactly the first rows of the full one. This
prefix-nesting property makes it possible to train a single language model that operates at several
vocabulary sizes, to condition it on which size is in use via a control token, and to deploy it at
any trained size by slicing rows from the embedding and output head. We pre-registered five claims
about this construction---fixing margins, seeds, contrasts, and an engineering stop rule before
spending compute---and trained 30 models (3.1M and 10.6M parameter bodies, 200M tokens each) to test
them.

The construction works exactly and its central premise does not. Slicing is numerically exact: a
sliced model reproduces the restricted full model's logits bit-for-bit across 76 checks, and removes
66\% of the deployed weights at unchanged latency. But a shared model trails a fixed-cap specialist
by $3.64\%$ bits-per-byte at 32k against a $1\%$ margin, and by $2.96\%$ at 8k against a $2\%$
margin. A $2\times2$ ablation separating the control token from the output restriction finds the
token worth $+0.07\%$ to $+0.13\%$ with all intervals crossing zero, while the output restriction
\emph{costs} $+0.47\%$ to $+1.19\%$; the two factors are substitutes rather than complements.

Two positive results survive. Multi-cap training confers genuine robustness to corrupted input: the
same checkpoint degrades 12.5--15.4 points less under typographical noise in its fine mode, and
\emph{outperforms} each fixed-cap specialist at that specialist's own vocabulary size. The
conditioning machinery earns no credit for this: a control trained across the same caps with no cap
token and no output restriction is equally robust, so training at several granularities regularises
against input noise on its own. And the per-cap penalty tracks each cap's share of training rows rather than any property of the conditioning
machinery, which yields a falsifiable prediction for future work.
\end{abstract}

\begin{IEEEkeywords}
tokenization, byte-pair encoding, nested vocabularies, language models, pre-registration,
negative results
\end{IEEEkeywords}

\section{Introduction}\label{sec:intro}

Subword vocabulary size is fixed at tokenizer construction and inherited by every model trained on
it. A coarse vocabulary yields short sequences and cheap inference but obscures character-level
structure; a fine vocabulary exposes orthography at the cost of longer sequences. Systems that need
both typically train two models.

Byte-level BPE \cite{b1}, \cite{b4} admits a structural shortcut. Because the merge table is
ordered, truncating it after $V_c$ merges produces a valid tokenizer whose identifiers are precisely
the first $V_c$ rows of the full vocabulary. Vocabularies at different sizes are therefore \emph{prefix-nested}: one embedding
matrix contains all of them. This suggests training a single model across several nested caps,
telling it which cap each sequence uses, and restricting its output distribution accordingly---so
that one artifact serves every granularity and can be reduced to any of them by deleting rows.

Nesting is a familiar idea, applied here to an unfamiliar axis. MatFormer nests feed-forward
widths and Matryoshka representation learning nests embedding dimensions, each training a model so
that a prefix of its hidden units stands alone \cite{b6}, \cite{b7}. In both, the nesting must be
optimised for: the property does not exist until it is trained in. The vocabulary is the case where
it exists already, because the merge table is ordered whether or not anyone exploits it---which is
what makes it worth asking whether a model can be handed the property for free.

The question is not feasibility, which is immediate, but cost. We pre-registered five claims,
fixed the evaluation protocol and margins before compute, and report what the experiment returned,
including the outcomes that contradict the design's motivating intuitions.

We work deliberately at 3.1M and 10.6M parameter bodies. The design this question needs---30 models,
paired seeds sharing initialisation and data order, a fully crossed $2\times2$, and a check of the
deployment path exact to the last bit---is affordable at that size and is not affordable at the
scales the construction would eventually serve, where the same comparisons would be run once and
read through seed noise. The trade is explicit: absolute performance for statistical resolution, and
a small-scale question decided rather than a large-scale one suggested. Section~\ref{sec:limits}
states what this therefore cannot establish.

\section{Method}

\subsection{Nested vocabularies}

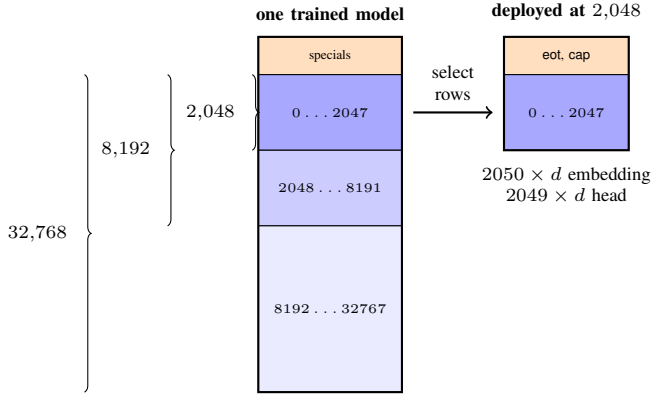
\begin{figure}[t]
\centering
\begin{tikzpicture}[x=1mm,y=1mm,font=\scriptsize]
  \def\W{19}
  \draw[fill=orange!25] (0,42) rectangle (\W,47);
  \draw[fill=blue!34]   (0,32) rectangle (\W,42);
  \draw[fill=blue!20]   (0,22) rectangle (\W,32);
  \draw[fill=blue!8]    (0,0)  rectangle (\W,22);
  \draw[thick] (0,0) rectangle (\W,47);
  \node[anchor=south] at (\W/2,48) {\textbf{one trained model}};
  \node at (\W/2,44.5) {\tiny specials};
  \node at (\W/2,37)   {\tiny $0\ldots2047$};
  \node at (\W/2,27)   {\tiny $2048\ldots8191$};
  \node at (\W/2,11)   {\tiny $8192\ldots32767$};
  \draw[decorate,decoration={brace,amplitude=2.5pt}] (-0.8,42) -- (-0.8,32)
    node[midway,left=3pt] {$2{,}048$};
  \draw[decorate,decoration={brace,amplitude=2.5pt}] (-12,42) -- (-12,22)
    node[midway,left=3pt] {$8{,}192$};
  \draw[decorate,decoration={brace,amplitude=2.5pt}] (-23,42) -- (-23,0)
    node[midway,left=3pt] {$32{,}768$};
  \draw[->,thick] (\W+1.5,37) -- (\W+12,37)
    node[midway,above=1pt,align=center] {select\\rows};
  \draw[fill=orange!25] (\W+13.5,42) rectangle (\W+30,47);
  \draw[fill=blue!34]   (\W+13.5,32) rectangle (\W+30,42);
  \draw[thick] (\W+13.5,32) rectangle (\W+30,47);
  \node[anchor=south] at (\W+21.75,48) {\textbf{deployed at $2{,}048$}};
  \node at (\W+21.75,44.5) {\tiny \textsf{eot}, \textsf{cap}};
  \node at (\W+21.75,37)   {\tiny $0\ldots2047$};
  \node[anchor=north,align=center] at (\W+21.75,31)
    {$2050\times d$ embedding\\$2049\times d$ head};
\end{tikzpicture}
\caption{Prefix nesting and slicing. Braces mark the three trained caps: each is a prefix of the
next, so deploying at a smaller cap is row selection rather than retraining or approximation. Bands
are not drawn to scale. Because the sliced model computes the same inner products over the same
rows, its logits equal the restricted full model's exactly---verified in
Section~\ref{sec:deploy}.}
\label{fig:nesting}
\end{figure}

A single byte-level BPE tokenizer is trained to $V_{\max}=32{,}768$ merge ranks. Caps
$\mathcal{K}=\{2{,}048;\ 8{,}192;\ 32{,}768\}$ are obtained by truncation. Because rank order is
preserved, a token identifier means the same thing at every cap that contains it, so one embedding
matrix contains every cap's as a prefix of its rows (Fig.~\ref{fig:nesting}).

\subsection{Conditioning}

Each training row draws a cap $k\sim p(k)=(0.2,0.3,0.5)$ over $\mathcal{K}$ with a recorded seed,
carries a control token naming $k$ at position 0, and has its output distribution restricted to the
$k$ lexical rows plus an end-of-text row. Conditions differ only in whether the control token is
informative and whether the restriction is applied, so a $2\times2$ design isolates each factor.
Paired seeds share initialization and data order, and the mixed-cap conditions receive byte-identical
rows differing only in the prefix identifier.

\subsection{Metric}

The primary metric is bits per byte (BPB): total negative log-likelihood in bits over predicted
targets, divided by the UTF-8 byte length of the decoded target text. Perplexity is never compared
across tokenizers, as tokens per byte differ by construction. Intervals are document-level block
bootstraps.

\section{Experimental Setup}\label{sec:setup}

Held-out sets were carved from FineWeb-Edu \cite{b5} before tokenizer training and
decontaminated: 6{,}473 training documents
sharing any 64-byte window with a held-out set were removed, after which the overlap check reports
zero shared windows at stride~1. Models are decoder-only transformers at two tiers---XS, a
3.1M-parameter body, and S, a 10.6M-parameter body---each trained with AdamW, 300 warmup steps,
cosine decay to 10\% of peak, batch $32\times1024$ positions, and a 200M-token budget fixed from
measured throughput before any experimental run. Embedding and output head are untied by design:
they are the two matrices a nested vocabulary slices, and tying them would confound input with
output capacity. A sensitivity run confirms tying is better at every cap, so the absolute figures
below are about one percent pessimistic---the comparisons are not, both sides being untied
(Section~\ref{sec:limits}). Every results row
carries the git commit, configuration hash, data hash, tokenizer hash, seed, and device that produced
it.

A pre-registered engineering stop rule---both of the first two shared-model seeds more than $3\%$
behind the coarse specialist---fired on the primary comparison and halted the larger tier, while the
factorial corners continued at the smaller tier so the negative result stayed interpretable.

\section{Results}

\subsection{Specialist equivalence (C1, C2)}

\begin{table}[htbp]
\caption{Equivalence tests against fixed-cap specialists, tier S, paired seeds}
\begin{center}
\begin{tabular}{lccc}
\toprule
\textbf{Comparison} & \textbf{Margin} & \textbf{Measured} & \textbf{Verdict} \\
\midrule
Shared vs.\ 32k specialist & $1\%$ & $+3.640\%$ & fails \\
 & & {\footnotesize $[+3.419,+3.861]$} & \\
Shared vs.\ 8k specialist & $2\%$ & $+2.958\%$ & fails \\
 & & {\footnotesize $[+2.846,+3.071]$} & \\
\bottomrule
\end{tabular}
\label{tab:equiv}
\end{center}
\end{table}

Both intervals lie wholly above their margins (Table~\ref{tab:equiv}). The gap is roughly 180 times
the seed-to-seed spread of the specialist itself, so this is not a resolution limit. The same
comparison at a $4\times$ smaller body gives $+3.64\%$---the larger-tier figure to three
digits---so the penalty is stable across scale.

\subsection{Explicit conditioning (C3)}

\begin{table*}[htbp]
\caption{The $2\times2$ ablation, three paired seeds, tier XS. Positive means higher BPB (worse).
Brackets are document-level block-bootstrap intervals.}
\begin{center}
\begin{tabular}{lccc}
\toprule
\textbf{Contrast} & \textbf{cap 2{,}048} & \textbf{cap 8{,}192} & \textbf{cap 32{,}768} \\
\midrule
Token, no restriction & $+0.071\%$ & $+0.125\%$ & $+0.076\%$ \\
 & {\footnotesize $[-0.408,+0.431]$} & {\footnotesize $[-0.103,+0.258]$}
 & {\footnotesize $[-0.047,+0.181]$} \\
Token, with restriction & $-0.102\%$ & $-0.093\%$ & $-0.180\%$ \\
 & {\footnotesize $[-0.301,+0.061]$} & {\footnotesize $[-0.208,+0.010]$}
 & {\footnotesize $[-0.367,-0.080]$} \\
Restriction, no token & $+1.192\%$ & $+0.608\%$ & $+0.472\%$ \\
 & {\footnotesize $[+1.080,+1.347]$} & {\footnotesize $[+0.476,+0.728]$}
 & {\footnotesize $[+0.260,+0.703]$} \\
Interaction & $-0.173\%$ & $-0.217\%$ & $-0.257\%$ \\
 & {\footnotesize $[-0.370,+0.107]$} & {\footnotesize $[-0.466,+0.023]$}
 & {\footnotesize $[-0.462,-0.047]$} \\
Total treatment & $+1.090\%$ & $+0.516\%$ & $+0.292\%$ \\
 & {\footnotesize $[+0.849,+1.280]$} & {\footnotesize $[+0.413,+0.648]$}
 & {\footnotesize $[+0.087,+0.609]$} \\
\bottomrule
\end{tabular}
\label{tab:2x2}
\end{center}
\end{table*}

Without the output restriction the control token does nothing measurable: every interval crosses
zero, and the point estimates carry the wrong sign for the claim (Table~\ref{tab:2x2}). Two results
that the design required to be reported regardless of sign cut against its motivating mechanism.
First, \textbf{output restriction costs rather than saves}: masking the head to the vocabulary
currently in use raises BPB by $0.47\%$ to $1.19\%$, every interval excluding zero, worst at the
finest cap. Second, \textbf{the token and the restriction are substitutes, not complements}: the
interaction is negative at all three caps, so the token is worth \emph{more} when the restriction is
already present---it recovers part of what masking costs---where the design expected the two to
reinforce. The direction is consistent across caps; the interval excludes zero at the coarsest cap
only, so the sign is better resolved than the magnitude.

The four corners order as $\text{blind} < \text{token-only} < \text{restriction-only} <
\text{both}$: the full treatment is $0.29\%$ to $1.09\%$ worse than telling the model nothing.

\subsection{Granularity dial (C4)}

\begin{table}[htbp]
\caption{Fine mode against coarse mode, and against specialists. Positive favours the fine mode.}
\begin{center}
\begin{tabular}{lcc}
\toprule
\textbf{Comparison} & \textbf{Orthographic} & \textbf{Typo} \\
 & \textbf{accuracy} & \textbf{robustness} \\
\midrule
Shared, fine vs.\ own 32k mode (XS) & $+0.021$ & $+0.154$ \\
Shared, fine vs.\ own 32k mode (S) & $-0.004$ & $+0.125$ \\
Shared vs.\ 8k specialist & $-0.083$ & $+0.042$ \\
Shared vs.\ 32k specialist & $-0.113$ & $+0.038$ \\
\midrule
Blind control, fine vs.\ 32k (XS) & $-0.017$ & $+0.161$ \\
\bottomrule
\end{tabular}
\label{tab:dial}
\end{center}
\end{table}

The claim splits by evaluation (Table~\ref{tab:dial}). On robustness to typographical corruption it
holds outright, and provides the strongest positive result in this work: the same checkpoint degrades
12.5--15.4 points less in its fine mode, and---uniquely in these experiments---\emph{outperforms}
each specialist at that specialist's own cap. On orthographic accuracy it fails: the dial is
marginal at the smaller tier, absent at the larger, and the shared model trails specialists by 8--11
accuracy points.

The blind control settles attribution. A model trained across caps but given no control token and no
restriction obtains the \emph{same} robustness ($+0.161$ against $+0.154$) and not the orthographic
dial ($-0.017$ against $+0.021$). Robustness is therefore a property of training across
granularities, available without any conditioning machinery; only the small orthographic effect is
attributable to the control token and the restriction.

\subsection{Deployment (C5)}\label{sec:deploy}

\begin{table}[htbp]
\caption{Deployment cost, tier S, fixed 10\,KB byte string, idle accelerator}
\begin{center}
\begin{tabular}{lccc}
\toprule
\textbf{Configuration} & \textbf{Weights} & \textbf{Prefill} & \textbf{Tokens} \\
 & \textbf{(MB)} & \textbf{(ms)} & \textbf{per byte} \\
\midrule
Full model & $71.58$ & $24.7$ & $0.2236$ \\
Sliced to cap 8{,}192 & $33.83$ & $24.9$ & $0.2560$ \\
Sliced to cap 2{,}048 & $24.39$ & $33.1$ & $0.3556$ \\
\bottomrule
\end{tabular}
\label{tab:cost}
\end{center}
\end{table}

Slicing is numerically exact: across 76 checks spanning every run and cap, the sliced model's logits
equal the restricted full model's with a worst-case absolute difference of $0.000\mathrm{e}{+}00$
against a pre-registered tolerance of $10^{-5}$. This is bit-identity rather than approximation,
because slicing selects rows and the restricted model computes the same inner products over them.

Consequently slicing is a free memory saving: reducing the artifact to cap 2{,}048 removes 66\% of
the weights at latency unchanged within measurement noise (Table~\ref{tab:cost}). What costs is the
granularity itself---running at cap 2{,}048 rather than 32{,}768 takes 34\% longer for the same text
because it requires 59\% more tokens. The deployment trade is not shared-model-versus-specialist but
granularity-versus-throughput.

\section{Discussion}

\subsection{The penalty is data dilution}\label{sec:dilution}

\begin{figure}[t]
\centering
\includegraphics[width=\columnwidth]{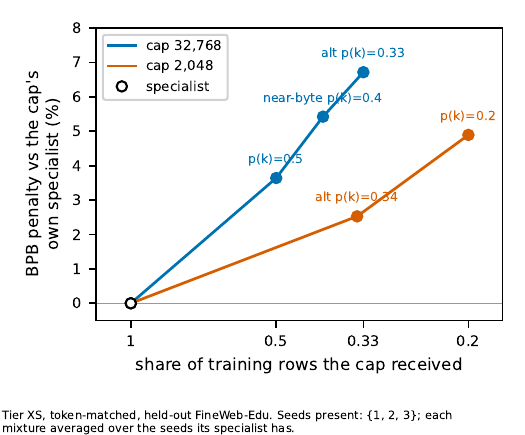}
\caption{The per-cap penalty against how much data the cap received. Each filled point is a
mixed-cap model charged against the fixed-cap specialist for its cap; the specialist gave that cap
every row it had and is therefore the zero. Both caps are monotone in the share. The specialists at
this tier were added outside the pre-registration after the stop rule fired, and the
alternative-$p(k)$ and near-byte mixtures are single-seed sensitivity runs. Cap 8{,}192 is absent
because its only specialist is at tier S, where the same comparison reads $+2.96\%$ at a share of
$0.30$.}
\label{fig:dilution}
\end{figure}

Three observations converge. The per-cap deficit tracks each cap's share of training rows
(Fig.~\ref{fig:dilution}): with shares $1.00$, $0.50$, and $0.33$ at cap 32{,}768, BPB is $1.3333$,
$1.3818$, and $1.4228$, a cost of $0.070$ then $0.099$ per natural-log unit of data---the same order
and monotone. The relation survives a sign flip: the alternative mixture gives cap 32{,}768 fewer
rows than the pre-registered one and is the worse model there ($+6.72\%$ against $+3.64\%$), and
gives cap 2{,}048 more rows than the pre-registered one and is the better model there ($+2.53\%$
against $+4.89\%$). The ordering follows the share, not the condition. Neither the control
token nor the output restriction adds data, and neither recovers the deficit. And the restriction,
which removes gradient signal from excluded rows without supplying any elsewhere, actively costs.

The prediction that follows is falsifiable and untested here: holding \emph{per-cap} tokens fixed
rather than total tokens should shrink the gap toward zero at proportionally greater compute. That,
rather than further attempts to make the label matter, is the experiment these results argue for.

\subsection{Two results of independent interest}

\textbf{Output restriction is not free.} Masking a model's output head to the vocabulary it is
currently using is intuitively harmless---the excluded rows cannot be correct---and is measurably
harmful, by up to $1.19\%$ BPB. Practitioners who restrict decoding to a sub-vocabulary during
training should expect to pay for it.

\textbf{Labels a model can infer are worth nothing.} Neither the vocabulary cap nor, in a second
setting, whether an input was produced by BPE-dropout ($-0.033\%$) changed loss measurably. Two
independent conditioning signals, both inferable from token statistics, both inert. A control token
occupies a position and a vocabulary row; these results suggest that cost is not repaid when the
model can recover the same information from the input.

\subsection{Multi-granularity training as implicit regularisation}

The robustness result and its control together describe an effect the conditioning machinery earns
no credit for. The shared model degrades 12.5--15.4 points less than a specialist under
typographical noise and beats every specialist at that specialist's own cap; the blind
control---same caps, same data, no cap token, no output restriction---degrades $0.161$ less against
the shared model's $0.154$. Whatever produces the robustness is already present before any of the
labelling is added.

A hypothesis, formed after seeing these numbers and so not pre-registered: training across caps is
segmentation-noise augmentation. The same text arrives under several segmentations, so no single
one can be relied on and the model must keep representations that survive being resegmented. A
typographical error is a resegmentation---the perturbation the training distribution already
contains. That is the mechanism BPE-dropout applies deliberately \cite{b2}, \cite{b3}, obtained here
as a side effect of varying the cap, which is consistent with the two techniques buying the same
robustness at different prices (Section~\ref{sec:dropout}).

The boundary matters as much as the effect. Multi-granularity training does not make the shared
model better; it makes it degrade less. The same checkpoint still pays $3.64\%$ BPB against a
specialist on clean text and still trails specialists by 8--11 points on orthographic accuracy. What
it buys is a flatter response to corrupted input, and it buys that whether or not the model is told
anything.

\subsection{Comparison with stochastic tokenization}\label{sec:dropout}

BPE-dropout \cite{b2} is the established method for making a model robust to segmentation, and
belongs to the same family as Kudo's subword regularization \cite{b3}. At the native cap it
costs $1.40\%$ against the specialist where multi-cap training costs $3.44\%$; asked to operate at
cap 2{,}048, the plain specialist scores $2.3748$ BPB, the dropout model $1.9271$, and the multi-cap
model $1.4609$. The two techniques purchase different amounts of the same robustness at different
prices, and which trade is preferable depends on how far from the native cap a deployment must
operate.

The rule that follows is a crossover rather than a winner. Near the native cap BPE-dropout is the
cheaper purchase, $1.40\%$ against $3.44\%$; the ordering reverses as the deployment moves away from
it, and by cap 2{,}048 the multi-cap model leads by $0.47$ BPB. Choose BPE-dropout when robustness
is wanted at one fixed granularity, and multi-cap training when a single artifact must span
granularities at inference time.

\section{Limitations}\label{sec:limits}

The scale is the choice argued for in Section~\ref{sec:intro}, and it bounds what can be claimed:
nothing here establishes behaviour at larger scale, and the vocabulary is a far larger share of
parameters than it would be in any model anyone would deploy. Roughly one
epoch of the pool, so models are far from converged. English-centric data with small code and
non-English held-out sets. A context confound that the byte-matched-context panel reduces
asymmetrically rather than removes: a fixed token window at a fine cap covers fewer bytes, and a
4{,}096-byte cap binds only above 4.0 bytes per token. One tokenizer, one pretokenizer, one data
mixture; tokenizer variance is not estimated. Tying embedding and head, which
Section~\ref{sec:setup} declines for the reason given there, is better at every cap with 42\% fewer
parameters, so the absolute figures here are approximately one percent pessimistic---though the
comparisons are unaffected, both sides being untied.

Measurement reproducibility differs by quantity: prefill latency repeats to $0.2\%$ and generated
bytes per second to about $5\%$, while the head matmul timed alone does not repeat within tolerance
and is reported flagged rather than quoted.

\section{Conclusion}

Nested byte-level vocabularies deliver on their engineering promise and not on their modelling
promise. A single model can be sliced to any trained vocabulary size with bit-identical outputs and a
66\% reduction in weights at no latency cost, which makes multi-granularity deployment practical. But
sharing capacity across caps costs $3.6\%$ BPB against a specialist, the visible cap label recovers
none of it, and restricting the output distribution to the active vocabulary makes matters worse.
What multi-cap training does buy is robustness---the shared model beats every specialist under
input corruption---and that benefit requires no conditioning machinery at all, arriving equally in a
model told nothing.

All code, configurations, pre-registration, results tables, and per-run manifests are available
at \url{https://github.com/unseen1980/captok}.

\section*{Acknowledgment}

The experimental campaign was executed autonomously under a pre-registration fixed before compute;
every amendment carries a dated rationale and a changelog entry in the repository.

\end{document}